\documentclass[12pt]{article}
\pdfoutput=1
\usepackage{graphicx,pict2e,latexsym,xcolor,amsmath,amssymb}
\usepackage[square,numbers]{natbib}
\newtheorem{defn}{Definition}
\newtheorem{theorem}{Theorem}
\newtheorem{lemma}{Lemma}

\def\M{{\mathcal M}}
\def\G{{\mathcal G}}

\def\X{{\mathcal X}}
\def\Y{{\mathcal Y}}

\newcommand{\iid}{\stackrel{iid}{\sim}}
\newcommand{\qed}{\hfill $\Box$}

\begin{document}
\bibliographystyle{plainnat}
\pagestyle{plain}

\title{\Large \bf Continuous Multidimensional Scaling}

\author{Michael W. Trosset\thanks{Department of Statistics, Indiana University.  
E-mail: {\tt mtrosset@iu.edu}} 
\and
Carey E. Priebe\thanks{Department of Applied Mathematics \& Statistics, Johns Hopkins University. 
E-mail: {\tt cep@jhu.edu}}
}

\date{\today}

\maketitle


\begin{abstract}
Multidimensional scaling (MDS) is the act of embedding proximity information about a set of $n$ objects in $d$-dimensional Euclidean space. As originally conceived by the psychometric community, MDS was concerned with embedding a fixed set of proximities associated with a fixed set of objects. Modern concerns, e.g., that arise in developing asymptotic theories for statistical inference on random graphs, more typically involve studying the limiting behavior of a sequence of proximities associated with an increasing set of objects. 
Here we are concerned with embedding dissimilarities by minimizing Kruskal's \cite{kruskal:1964a} raw stress criterion.
Standard results from the theory of point-to-set maps can be used to establish that, if $n$ is fixed and a sequence of dissimilarity matrices converges, then the limit of their embedded structures is the embedded structure of the limiting dissimilarity matrix. But what if $n$ increases? It then becomes necessary to reformulate MDS so that the entire sequence of embedding problems can be viewed as a sequence of optimization problems in a fixed space.  We present such a reformulation, {\em continuous MDS}.  Within the continuous MDS framework, we derive two $L^p$ consistency results, one for embedding without constraints on the configuration, the other for embedding subject to {\em approximate Lipschitz constraints}\/ that encourage smoothness of the embedding function.  The latter approach, {\em Approximate Lipschitz Embedding}\/ (ALE) is new.  Finally, we demonstrate that embedded structures produced by ALE can be interpolated in a way that results in uniform convergence.  
\end{abstract}


\newpage

\tableofcontents

\newpage


\section{Introduction}

We endeavor to fill a gap in the current literature on embedding pairwise dissimilarity data in Euclidean space.  Whereas the vast literature on multidimensional scaling (MDS) is almost entirely concerned with embedding dissimilarity data for a fixed number of objects, various modern applications are concerned with situations in which the number of objects tends to infinity.  For example:
\begin{itemize}

\item  Manifold learning studies the recovery of data manifolds.  Much of the current theory offers guarantees that hold asymptotically, as the manifold is sampled more and more extensively.  If recovery means representation in Euclidean space, then it is natural to inquire how the representations behave asymptotically.

\item  Network science frequently studies the behavior of graphs with increasing numbers of vertices.  Upon constructing the matrix of pairwise dissimilarities between the vertices of a graph, the problem of embedding that graph in Euclidean space (here referred to as {\em graph embedding}, although the same phrase has also been used in other contexts) reduces to the problem of MDS.  Again, it is natural to inquire how the Euclidean representations of the graphs behave asymptotically.

\end{itemize}

The preceding examples coalesce in the manifold learning technique known as Isomap \cite{isomap:2000}, the study of which motivated the present inquiry.  Isomap posits data that lie on a connected compact Riemannian manifold and consists of three distinct steps: (1) constructing a graph that summarizes the local structure of the data; (2) computing the shortest path distances between the vertices; and (3) embedding the shortest path distances by MDS.  Several researchers 
\cite{Bernstein&etal:2000,mwt:isomap}
have studied the convergence of the shortest path distances to the corresponding Riemannian distances as sampling of the manifold increases.  Here, we develop tools for examining the asymptotic behavior of the pairwise Euclidean distances in the embedded configuration.

The development that follows assumes that embedding is accomplished by finding a global minimizer of Kruskal's \cite{kruskal:1964a} raw stress criterion.  Similar results should be possible for other well-behaved embedding criteria, e.g., classical MDS \cite{torgerson:1952}.  We ignore the practical difficulty of finding global minimizers of this nonconvex objective function, but note that the raw stress criterion is usually minimized by numerical algorithms that are only guaranteed to find local solutions.  Computational issues are discussed in an appendix.

We apply elementary results from the theory of point-to-set maps, following the expositions in \cite{Hogan:1973,Anisiu:1981} and summarized in Section~\ref{maps}.  To illustrate our approach, and to identify some key difficulties in developing a suitable asymptotic theory, Section~\ref{fixed} examines the familiar case of a fixed number ($n$) of points.  In this case, one can formulate the problem of minimizing a fixed raw stress criterion as the problem of minimizing the (squared) Frobenius norm of the difference between a fixed $n \times n$ dissimilarity matrix and elements of the cone of $n \times n$ matrices of pairwise Euclidean distances.
In this formulation, the distances are the decision variables and the dissimilarities parametrize a family of optimization problems.
We establish that, if a sequence of dissimilarity matrices converges to a limiting dissimilarity matrix, then (a) any sequence of solutions contains an accumulation point, and (b) any accumulation point solves the limiting optimization problem.

Although conceptually identical, the various raw stress criteria associated with varying $n$ are formally distinct.  In Section~\ref{compact}, we replace a finite set of objects with a probability measure on a compact set of objects, dissimilarity matrices with dissimilarity functions, distance matrices with distance functions, and finite sums with integration with respect to probability measures.  The resulting framework of {\em continuous multidimensional scaling}\/ permits application of point-to-set map theory as $n$ varies.  Because the feasible sets of the optimization problems are now of infinite dimension, the arguments are more subtle and one must be careful to use the appropriate topology.
We establish an $L^p$ consistency result for embedding by unconstrained minimization of raw stress.  If the dissimilarity functions converge as $n \rightarrow \infty$, then (a) any sequence of solutions contains an accumulation point, and (b) any accumulation point solves a limiting problem.

We investigate the possibility of uniform convergence in Section~\ref{ApproxLipCon}.  First, we introduce a new MDS technique, {\em Approximate Lipschitz Embedding}\/ (ALE), that embeds dissimilarity matrices by minimizing raw stress subject to {\em approximate Lipschitz constraints}\/ that encourage smoothness of the embedding function.  Again, computational issues are discussed in an appendix.  Next we establish an $L^p$ consistency result for ALE.  Finally we demonstrate that it is possible to interpolate configurations constructed by ALE in such a way that the interpolating functions converge uniformly.

Section~\ref{disc} concludes.

\section{Point-to-Set Maps}
\label{maps}

Our brief exposition draws from \cite{Hogan:1973} and \cite{Anisiu:1981}.
Let $\X$ and $\Y$ denote complete metric spaces.  Notice that every such space has a countable basis.  A point-to-set map from $\X$ to $\Y$ assigns a subset of $\Y$ to each point in $\X$.

\begin{defn}
Let $\Omega : \X \multimap \Y$ denote a point-to-set map from $\X$ to $\Y$,
and suppose that $\bar{x} \in \X$.
\begin{itemize}

\item Suppose that, for every sequence $\{ x_k \} \subset \X$ that converges to $\bar{x}$ and every $\bar{y} \in \Omega(\bar{x})$, there exists a natural number $\ell$ and a sequence $\{ y_k \} \subset \Y$ that converges to $\bar{y}$ for which $y_k \in \Omega(x_k)$ for $k \geq \ell$.  Then $\Omega$ is {\em open}\/ at $\bar{x}$.

\item Suppose that, for every sequence $\{ x_k \} \subset \X$ that converges to $\bar{x}$, the limit of every convergent sequence $\{ y_k \in \Omega(x_k) \}$ lies in $\Omega(\bar{x})$.  Then $\Omega$ is {\em closed}\/ at $\bar{x}$. 

\end{itemize}
If $\Omega$ is both open and closed at $\bar{x}$, then $\Omega$ is {\em continuous}\/ at $\bar{x}$.  If $\Omega$ is open (respectively closed, continuous) at every $\bar{x} \in \X$, then $\Omega$ is open (respectively closed, continuous) on $\X$.
\label{def:map}
\end{defn}

The preceding definition of an open point-to-set map is equivalent to the second (III.1.1$^\prime$) of two definitions of a lower semicontinuous (l.s.c.) point-to-set map in \cite{Anisiu:1981}.  If both $\X$ and $Y$ have countable bases, then the two definitions are the same \cite[Corollary~III.1.1]{Anisiu:1981}.  Theorems III.1.5 and III.1.6 in \cite{Anisiu:1981} provide characterizations of lower semicontinuity;
in particular, Theorem~III.1.6 states that $\Omega : \X \multimap \Y$ is l.s.c.\ if and only if, for any closed $H \subset \Y$, the set
$\{ x \in \X : \Omega(x) \subset H \}$ is closed in $\X$.

Furthermore, the preceding definition of a closed point-to-set map is equivalent to the second (III.3.1$^\prime$) of two definitions in \cite{Anisiu:1981}.  Again, if both $\X$ and $Y$ have countable bases, then the two definitions are the same \cite[Theorem~III.3.3]{Anisiu:1981}.  Theorem~III.3.1 states that, if $\Omega : \X \multimap \Y$ is closed in the sense of the first definition (III.3.1), then each $\Omega(x)$ is a closed subset of $\Y$.

Let $C \subset \Y$ be any closed subset of $\Y$, and
define a constant point-to-set map
$\Omega : \X \multimap \Y$ by $\Omega(x) = C$ for every $x \in \X$.
It is evident that $\Omega$ is both open and closed, hence continuous.

The intersection $\Omega_1 \cap \Omega_2$ of point-to-set maps $\Omega_1, \Omega_2 : \X \multimap \Y$ is defined by 
\[
\left( \Omega_1 \cap \Omega_2 \right) (x) = \Omega_1(x) \cap \Omega_2(x).
\]
The intersection of closed point-to-set maps is necessarily closed
\cite[Theorem~III.3.6]{Anisiu:1981}, 
whereas the intersection of open point-to-set maps is not necessarily open \cite[Theorem~III.5.3]{Anisiu:1981}.
  
The theory of point-to-set maps has been widely used to study the convergence properties of algorithms for mathematical programming, especially in the pioneering work of Zangwill \cite{Zangwill:1969}.
Continuing to follow \cite{Hogan:1973},
let $\sigma: \X \times \Y \rightarrow [-\infty,+\infty]$ and define $f: \X \times \Y \rightarrow [-\infty,+\infty]$ by $f(x,y)=-\sigma(x,y)$.  We will interpret $\sigma(x,\cdot)$ as an objective function to be minimized in $y$, so that maximizing $f(x,\cdot)$ in $y$ is equivalent to minimizing $\sigma(x,\cdot)$ in $y$.

Let $\Omega(x) \subset \Y$ denote the set of feasible $y$ when minimizing $\sigma(x,\cdot)$ or, equivalently, maximizing $f(x,\cdot)$.  The {\em supremal value function}\/ $v : \X \rightarrow [-\infty,+\infty]$ is defined by
\[
v(x) = \sup \left\{ f(x,y) : y \in \Omega(x) \right\} 
 = - \inf \left\{ \sigma(x,y) : y \in \Omega(x) \right\} 
\]
and the set of global maximizers of $f(x,\cdot)$ (equivalently, global minimizers of $\sigma(x,\cdot)$) is defined by
\[
\mbox{\tt Min}(x) = \left\{ y \in \Omega(x) : f(x,y) \geq v(x) \right\}
 = \left\{ y \in \Omega(x) : \sigma(x,y) \leq -v(x) \right\}.
\]
We will rely on the following property of the point-to-set map $\mbox{\tt Min} : \X \multimap \Y$:
\begin{theorem}[{\cite[Theorem~8]{Hogan:1973}}]
If $\Omega$ is continuous at the point $\bar{x} \in \X$ and $f$ (equivalently, $\sigma$) is continuous on the set $\bar{x} \times \Omega(\bar{x})$, then $\mbox{\tt Min}$ is closed at $\bar{x}$. 
\label{thm:Hogan8}
\end{theorem}
In essence, Theorem~\ref{thm:Hogan8} states conditions under which the optimization problem
\begin{equation}
\begin{array}{ll}
\mbox{minimize} & \sigma(x,y) \\ 
\mbox{subject to} & y \in \Omega(x)
\end{array}
\label{pr:Hogan}
\end{equation}
behaves nicely under perturbations in $x$, the parameters that define it.  More precisely, it states conditions ensuring that, if $x_k \rightarrow \bar{x}$ defines a sequence of problems and $y_k \rightarrow \bar{y}$ is a convergent sequence of solutions, then the limit of solutions is a solution of the limiting problem.

\section{Embedding a Fixed Number of Points}
\label{fixed}

For pedagogical reasons, we first consider the problem of embedding a fixed number of points.
Fix $n$, let $\Delta$ denote an $n \times n$ dissimilarity matrix with entries $\Delta(i,j)$, let $z_1,\ldots,z_n \in \Re^d$ denote an embedding of $\Delta$, and write
\[
Z = [ \begin{array}{c|c|c}
z_1 & \cdots & z_n 
\end{array} ]^t.
\]
The traditional weighted raw stress criterion is any scalar multiple of
\begin{equation}
\sigma_n(\Delta,Z) =  \sum_{i,j=1}^n  w_{ij}
\left[ \left\| z_i-z_j \right\| - \Delta(i,j) \right]^2,
\label{eq:rawstress}
\end{equation}
where the $w_{ij}=w_{ji}$ are finite nonnegative weights.
A popular approach to (metric) MDS is to fix $\Delta$, then find $Z$ that minimizes $\sigma_n$.  See the Appendix for a brief discussion of some numerical algorithms that attempt to do so.

Here we study the unweighted case, i.e., the special case in which each weight has the same value.
We scale the weights so that each
$w_{ij} = 1/n^2$ (instead of the conventional $w_{ij}=1$) in order to suggest integration of an error criterion with respect to a probability measure.
We say that an $n \times n$ matrix $D=[d_{ij}]$ is EDM-1 if and only if there exists a dimension $p$ and points $z_1,\ldots,z_n \in \Re^p$ such that $d_{ij} = \| z_i-z_j \|$.  The smallest such $p$ is the embedding dimension of $D$.
If $D$ denotes an $n \times n$ EDM-1 matrix of embedding dimension $\leq d$, then we can rewrite $\sigma_n$ as
\[
\sigma_n(\Delta,D) = \frac{1}{n^2} \| D-\Delta \|_F^2,
\]
where $\| \cdot \|_F$ is the Frobenius norm, and formulate the problem of minimizing the raw stress criterion as the problem of minimizing $\sigma_n$ as $D$ varies in $\Y_n$, the closed cone of all $n \times n$ EDM-1 matrices of embedding dimension $\leq d$.  Note that $\Y_n$ is a complete metric space in the topology of the Frobenius norm.

Now we allow $\Delta$ to vary. Let ${\mathcal X}_n$ denote the closed cone of all $n \times n$ dissimilarity matrices, which is also a complete metric space in the topology of the Frobenius norm.
Let $\Omega$ denote the constant point-to-set map from $\X_n$ to $\Y_n$ defined by $\Omega(\Delta)=\Y_n$. 
Let {\tt Min} denote the point-to-set map from $\X_n$ to $\Y_n$ defined by
\[
{\tt Min}(\Delta) = \left\{ D \in {\mathcal Y}_n :
\sigma_n(\Delta,D) \leq \inf_{D \in {\mathcal Y}_n} \sigma_n(\Delta,D) \right\},
\]
the set of globally minimizing EDM-1 matrices for the dissimilarity matrix $\Delta$.

Let $\{ \Delta_k \} \subset \X_n$ denote a sequence of dissimilarity matrices that converges to a dissimilarity matrix $\Delta_\infty \in \X_n$ in the sense that
\[
\lim_{k \rightarrow \infty} 
\left\| \Delta_k  - \Delta_\infty \right\|_F = 0.
\]
We study the behavior of ${\tt Min}(\Delta_k)$ as $k \rightarrow \infty$.  Specifically, we use the theory of point-to-set maps to demonstrate the following result:
\begin{theorem}
Fix $n$ and suppose that $\{ \Delta_k \}$ converges to $\Delta_\infty$.  Then
\begin{itemize}
\item[(a)] any sequence of $D_k \in {\tt Min}(\Delta_k)$ has an accumulation point, and
\item[(b)] if $D_\infty$ is an accumulation point of the sequence $\{ D_k \in {\tt Min}(\Delta_k) \}$, \\ then $D_\infty \in {\tt Min}(\Delta_\infty)$.
\end{itemize}
\label{thm:fixed}
\end{theorem}

\subparagraph{Proof}

Let ${\mathcal X}_n(\delta)$ denote the closed and bounded set of $n \times n$ dissimilarity matrices with Frobenius norm no greater than $\delta$.
Because $\| \Delta_k - \Delta_\infty \|_F \rightarrow 0$, there exists $\delta < \infty$ such that $\Delta_\infty,\Delta_1,\Delta_2,\ldots \in {\mathcal X}(\delta)$.

The set ${\tt Min}(\Delta)$ is a subset of the set of critical points of $\sigma_n(\Delta,\cdot)$.
Applying Theorem~5 in \cite{malone&etal:start} in the case of the raw stress criterion ($r=1$), all such critical points lie on a sphere in $\Re^{n \times n}$, centered at $\Delta/2$ with radius $\| \Delta/2 \|_F$.  It follows that, if $D \in {\tt Min}(\Delta)$, then $\| D \|_F \leq \| \Delta \|_F$.  If $\Delta \in {\mathcal X}_n(\delta)$, then $\| D \|_F \leq \delta$.

Let $\Y_n(\delta)$ denote the set of $n \times n$ EDM-1 matrices with embedding dimension $d$ and Frobenius norm no greater than $\delta$.  Because the finite-dimensional set $\Y_n(\delta) \subset \Re^{n \times n}$ is closed and bounded, it must be compact; hence the sequence $\{ D_k \} \subset \Y_n(\delta)$ must contain a convergent subsequence,
establishing (a).

Now suppose that $\Delta_k \rightarrow \Delta_\infty$ is such that
$D_k \rightarrow D_\infty$ as $k \rightarrow \infty$.
Observe that
\begin{itemize}

\item $\Omega$ is both an open and a closed point-to-set map, hence a continuous point-to-set map; and

\item for any $\Delta \in \X_n$, $\sigma_n$ is continuous on the set $\Delta \times \Y_n$.

\end{itemize}
Hence, by Theorem~\ref{thm:Hogan8}, ${\tt Min}$ is closed at $\Delta$.  As $\Delta$ is arbitrary, ${\tt Min}$ is a closed point-to-set map, i.e.,
\begin{eqnarray*}
\Delta_k \rightarrow \Delta_\infty, \;
D_k \in {\tt Min} \left( \Delta_k \right), \;
D_k \rightarrow D_\infty & \mbox{ entails } &
D_\infty \in {\tt Min} \left( \Delta_\infty \right),
\end{eqnarray*}
which is (b).  \qed

\bigskip

Theorem~\ref{thm:fixed} states that any accumulation point of embeddings obtained by minimizing a sequence of raw stress criteria is the embedding obtained by minimizing a limiting raw stress criterion.
Notice that the proof of (a) relies on the fact that closed and bounded implies compact in finite-dimensional Euclidean space.
Notice that the proof of (b) relies on our ability to formulate all of the relevant embedding problems as special cases of a general optimization problem of the form (\ref{pr:Hogan}), with fixed complete metric spaces $\X$ and $\Y$ and fixed functional $\sigma$.
Finally, notice that Theorem~\ref{thm:fixed} states that $D_\infty$ is an accumulation point of $\{ D_k \}$ in the topology of $L^2$ convergence.  Because $\{ D_k \}$ exists in finite-dimensional Euclidean space, $L^2$ convergence is topologically equivalent to uniform ($L^\infty$) convergence.  In Section~\ref{compact}, the analogous sequences will exist in an infinite-dimensional function space in which $L^2$ convergence does not imply uniform convergence.

\section{Embedding a Compact Set of Points}
\label{compact}

It is not obvious how to apply the techniques deployed in Section~\ref{fixed} to situations in which the number of points varies.  
Nor is it clear what kind of objects are suitable limits of $n \times n$ Euclidean distance matrices as $n$ tends to infinity.
To proceed, we must construct complete metric spaces and functional $(\X,\Y,\sigma)$ that include the $(\X_n,\Y_n,\sigma_n)$ of Section~\ref{fixed} as special cases.  The key to our construction lies in replacing finite sets of points with compact sets of points, then introducing probability measures that permit extraction of finite subsets of points.

Let ${\mathcal M}$ denote a compact metric space, e.g., a Riemannian manifold that is compact in the (Riemannian) metric topology.  Let ${\mathcal B}$ denote the Borel sigma-field generated by the open subsets of ${\mathcal M}$.  Let ${\mathcal P}$ denote the space of probability measures $P$ on the measurable space $({\mathcal M},{\mathcal B})$, topologized by weak convergence.  Because ${\mathcal M}$ is separable, the topology of weak convergence in ${\mathcal P}$ is equivalent to the topology of the Prohorov metric, with respect to which ${\mathcal P}$ is a complete metric space.  Furthermore,
because ${\mathcal M}$ is compact, ${\mathcal P}$ is compact in the topology of weak convergence 
\cite[Proposition~5.3]{GaansProbabilityMO}.

Let $\Delta : {\mathcal M} \times {\mathcal M} \rightarrow \Re$ be a Borel-measurable dissimilarity function on ${\mathcal M}$, i.e., $\Delta$ satisfies
$\Delta \left( m_1,m_1 \right) = 0$,
$\Delta \left( m_1,m_2 \right) \geq 0$, and
$\Delta \left( m_1,m_2 \right) = \Delta \left( m_2,m_1 \right)$
for any $m_1,m_2 \in {\mathcal M}$.
Let ${\mathcal D}$ denote the cone of all such $\Delta$.
Notice that ${\mathcal D}$ is closed, both in the topology of pointwise convergence and in the topology of $L^p$ convergence for any $p \in [1,\infty)$.  It is a complete metric space with respect to the latter.

Let ${\mathcal X} = {\mathcal D} \times {\mathcal P}$, the Cartesian product of ${\mathcal D}$ and ${\mathcal P}$, which is a complete metric space with respect to the product metric of the $L^p$ metric on ${\mathcal D}$ and the Prohorov metric on ${\mathcal P}$.
For $\delta >0$, let ${\mathcal X}(\delta)$ denote the closed subset of ${\mathcal X}$ for which 
\[
\sup_{(m_1,m_2) \in {\mathcal M} \times {\mathcal M}} 
\Delta \left( m_1,m_2 \right) \leq \delta.
\] 

Let ${\tt mds} : {\mathcal M} \rightarrow \Re^d$ denote a Borel-measurable embedding function and define the {\em continuous raw stress criterion}\/ by
\[
\sigma \left( (\Delta, P), {\tt mds} \right) =
\int_{\mathcal M} \int_{\mathcal M}
\left[ \left\| {\tt mds} \left( m_1 \right) -
{\tt mds} \left( m_2 \right) \right\| 
- \Delta \left( m_1,m_2 \right) \right]^2
P \left( dm_1 \right) P \left( dm_2 \right) .
\]
Notice that, if $P$ is a discrete probability measure supported on $n$ points, then we recover the traditional weighted raw stress criterion.

Observe that any function $D : {\mathcal M} \times {\mathcal M} \rightarrow \Re$ defined by
\begin{equation}
D \left( m_1,m_2 \right) = \left\| {\tt mds} \left( m_1 \right) -
{\tt mds} \left( m_2 \right) \right\|
\label{eq:D}
\end{equation}
is a pseudometric on ${\mathcal M}$.  We will refer to pseudometrics of this form as $d$-dimensional Euclidean pseudometrics.  Let ${\mathcal Y}$ denote the cone of all $d$-dimensional Euclidean pseudometrics.  Like ${\mathcal D}$, ${\mathcal Y}$ is closed, both in the topology of pointwise convergence and in the topology of $L^p$ convergence for any $p \in [1,\infty)$, and it is a complete metric space with respect to the latter.

Finally, rewrite the continuous raw stress criterion as
\[
\sigma \left( (\Delta, P), D \right) =
\left\| D-\Delta \right\|_P^2.
\]
Let $\Omega$ denote the constant point-to-set map from $\X$ to $\Y$ defined by $\Omega(\Delta,P)=\Y$, and note that $\Omega$ is continuous in the sense of Definition~\ref{def:map}.
Let 
\[
{\tt Min}(\Delta,P) = \left\{ D \in {\mathcal Y} :
\sigma((\Delta,P),D) \leq \inf_{D \in {\mathcal Y}} \sigma((\Delta,P),D) \right\},
\]
the set of globally minimizing $y = D \in \Y$ for the fixed pair $x = (\Delta,P) \in \X$.

Now suppose that $m_1,m_2,\ldots \iid P_\infty$.  Let $\hat{P}_n$ denote the empirical probability measure of $m_1,\ldots,m_n$.  It is well-known that $\hat{P}_n$ converges weakly to $P_\infty$.  Let $\Delta_n$ denote uniformly bounded dissimilarity functions that converge pointwise to a dissimilarity function $\Delta_\infty$ as $n \rightarrow \infty$.  For example, $\Delta_n$ might denote shortest path distance on a suitable graph constructed from $m_1,\ldots,m_n$ and $\Delta_\infty$ might denote Riemannian distance on a Riemannian manifold ${\mathcal M}$.  We study the behavior of ${\tt Min}(\Delta_n,\hat{P}_n)$ as $n \rightarrow \infty$.  To do so, we require several lemmas.

\begin{lemma}
Fix $0 \leq \delta_1,\delta_2 \leq \delta$ and $w_1,w_2>0$.
Set $\rho^2 = \max \{ w_1,w_2 \}/\min \{ w_1,w_2 \}$.
Suppose that $z_1,z_2,z \in \Re^2$ satisfy
$\| z_1-z_2 \| > 3\delta (1+\rho^2)$
and $\| z_i-z \| \leq \| z_1-z_2 \|$.
Let $u = (z_1-z_2)/\| z_1-z_2 \|$.
Then $0<\epsilon \leq \delta^{1/2}$ implies that $z_1^\prime = z_1-\epsilon u$ and $z_2^\prime = z_2+\epsilon u$ satisfy
\begin{equation}
\sum_{i=1}^2
w_i \left( \left\| z_i-z \right\| -\delta_i \right)^2 -
\sum_{i=1}^2
w_i \left( \left\| z_i^\prime-z \right\| -\delta_i \right)^2 \geq
\left( w_1+w_2 \right) \epsilon^2.
\label{eq:decrease}
\end{equation}
\label{lm:decrease}
\end{lemma}

\subparagraph{Proof}
Without loss of generality, choose the coordinate system in which $z_i=( \pm c,0)$ and $z=(a,b)$ with $a,b>0$.  If necessary, relabel the $z_i$ (and the corresponding $\delta_i$ and $w_i$) so that $z_1=(c,0)$ and $z_2=(-c,0)$.  Because $c = \| z_1-z_2 \|/2 > \delta$, it is impossible for both $\| z_i-z \| \leq \delta_i$ to obtain.

Using Taylor's Theorem, write
\[
f(\epsilon) = \left[ (c-a-\epsilon)^2 + b^2 \right]^{1/2} 
 = f(0) + f^\prime(0) \epsilon + R(\xi) \epsilon^2 
 = f(0) -\frac{c-a}{f(0)} \epsilon + 
 \frac{b^2}{2 f(\xi)^3} \epsilon^2,
\]
where $\xi \in [0,\epsilon]$.
Noting that $R(\xi) \geq 0$ and $0 < c-a \leq f(0)$, conclude that
\begin{eqnarray*}
\lefteqn{\left( \left\| z_1-z \right\| -\delta_1 \right)^2 -
\left( \left\| z_1^\prime-z \right\| -\delta_1 \right)^2} \\
 & = &
(c-a)^2+b^2 - 2 \delta_1 f(0) + \delta_1^2
- (c-a-\epsilon)^2-b^2 + 2 \delta_1 f(\epsilon) -\delta_1^2 \\
 & = &
2(c-a) \epsilon - \epsilon^2 + 2 \delta_1
\left[ -\frac{c-a}{f(0)} \epsilon + R(\xi) \epsilon^2 \right] \\
 & \geq &
2(c-a) \epsilon - 2 \delta_1 \frac{c-a}{f(0)} \epsilon -
\epsilon^2  \\
 & \geq & 2 \left( c-a-\delta_1 \right) \epsilon -
\epsilon^2 .
\end{eqnarray*}
Similarly, write
\[
g(\epsilon) = \left[ (c+a-\epsilon)^2 + b^2 \right]^{1/2} 
 = g(0) + g^\prime(0) \epsilon + S(\xi) \epsilon^2  
 = g(0) -\frac{c+a}{g(0)} \epsilon + \frac{b^2}{2 g(\xi)^3} \epsilon^2,
\]
where $\xi \in [0,\epsilon]$.
Noting that $S(\xi) \geq 0$ and $0 < c+a \leq g(0)$, conclude that
\begin{eqnarray*}
\lefteqn{\left( \left\| z_2-z \right\| -\delta_2 \right)^2 -
\left( \left\| z_2^\prime-z \right\| -\delta_2 \right)^2} \\
 & = &
(c+a)^2+b^2 - 2 \delta_2 g(0) + \delta_2^2
- (c+a-\epsilon)^2-b^2 + 2 \delta_2 g(\epsilon) -\delta_2^2 \\
 & = &
2(c+a) \epsilon - \epsilon^2  + 2 \delta_2
\left[ -\frac{c+a}{g(0)} \epsilon + S(\xi) \epsilon^2 \right] \\
 & \geq &
2(c+a) \epsilon - 2 \delta_2 \frac{c+a}{g(0)} \epsilon -
\epsilon^2  \\
 & \geq & 2 \left( c+a-\delta_2 \right) \epsilon - \epsilon^2 .
\end{eqnarray*}

Notice that
\begin{eqnarray*}
c-a-\delta_1 \geq -\delta & \mbox{and} &
c+a-\delta_2 \geq c + (c-\delta) -\delta = 2c - 2\delta ;
\end{eqnarray*}
hence,
\begin{eqnarray*}
w_1 \left( c-a-\delta_1 \right) +
w_2 \left( c+a-\delta_2 \right) & \geq &
w_1 (-\delta) + w_2 (2c-2\delta) \\
 & = &
2w_2 c - \left( w_1 + 2w_2 \right) \delta \\
 & \geq &  \left[ 3w_2(1+\rho^2) - \left( w_1 + 2w_2 \right) \right] \delta \\
 & \geq & \left[ 3w_2 \left( w_1+w_2 \right)/w_2 - w_1 -2w_2 \right] \delta \\
 & = & \left( 2w_1+w_2 \right) \delta \geq \left( w_1+w_2 \right) \delta.
\end{eqnarray*}
If $\epsilon^2 \leq \delta$, then
\begin{eqnarray*}
\lefteqn{w_1 \left[ 2 \left( c-a-\delta_1 \right) \epsilon -
\epsilon^2 \right] + w_2 \left[ 2 \left( c+a-\delta_2 \right) \epsilon - \epsilon^2 \right]} \\ & = &
2 \left[ w_1 \left( c-a-\delta_1 \right) +
w_2 \left( c+a-\delta_2 \right) \right] - \left( w_1+w_2 \right) \epsilon^2 \\
 & \geq & 2 \left( w_1+w_2 \right) \delta - \left( w_1+w_2 \right) \epsilon^2 \geq \left( w_1+w_2 \right) \epsilon^2 .
\end{eqnarray*}

\qed

\begin{lemma}
Let $\Delta : {\mathcal M} \times {\mathcal M} \rightarrow \Re$ be a dissimilarity function with
\[
\sup_{(m_1,m_2) \in {\mathcal M} \times {\mathcal M}} 
\Delta \left( m_1,m_2 \right) \leq \delta < \infty.
\]
Let $\hat{P}_n$ be an empirical measure on $({\mathcal M},{\mathcal B})$, i.e., a discrete probability measure that assigns probability $1/n$ to each of $n$ points in ${\mathcal M}$.
If $D \in {\tt Min}(\Delta,\hat{P}_n)$, then
\[
\sup_{(m_1,m_2) \in {\mathcal M} \times {\mathcal M}} 
D \left( m_1,m_2 \right) \leq 6\delta .
\]
\label{lm:Dbounded}
\end{lemma}

\subparagraph{Proof}
Let $m_1,\ldots,m_n \in {\mathcal M}$ denote the points on which $\hat{P}_n$ concentrates and let $z_i = \mbox{\tt mds}(m_i) \in \Re^d$.
Assume that the points have been indexed so that
$\| z_1-z_2 \| \geq \| z_i-z_j \|$
for every $i,j=1,\ldots,n$.
The proof is by contradiction.  We show that, if $\| z_1-z_2 \| > 6 \delta$, then $\sigma$ can be decreased by modifying $z_1$ and $z_2$.

In this setting, the objective function $\sigma$ simplifies to $\sigma_n$ with $w_{ij} = 1/n^2$.  Choose any $\epsilon \in (0,\delta^{1/2})$ and apply Lemma~\ref{lm:decrease} with each $z=z_1,\ldots,z_n$.  With each application, $\rho^2 = (1/n^2)/(1/n^2) = 1$, yielding
\begin{eqnarray*}
\lefteqn{\sigma_n (\Delta,Z) - \sigma_n \left( \Delta,Z^\prime \right)} \\ & = &
2 \sum_{i=1}^2 \sum_{j=i+1}^n \frac{1}{n^2} \left[ 
\left( \left\| z_i-z_j \right\| - \Delta(i,j) \right)^2 -
\left( \left\| z_i^\prime-z_j \right\| - \Delta(i,j) \right)^2
\right] \\ & \geq &
2 \sum_{i=1}^2 \sum_{j=i+1}^n \left( \frac{1}{n^2}+\frac{1}{n^2} \right) \epsilon^2 =
\frac{4(4n-6)}{n^2} \epsilon^2 > 0.
\end{eqnarray*}
Thus, the embedding $z_1,\ldots,z_n$ is suboptimal.
\qed

\bigskip

\begin{lemma}[Kehoe\cite{Kehoe:2019}]
Let ${\mathcal Y}(\gamma)$ denote the set of Euclidean pseudometrics on ${\mathcal M}$ that are bounded by $\gamma \in (0, \infty)$.  
Then any sequence $D_1,D_2,\ldots \in {\mathcal Y}(\gamma)$ has a subsequence that is pointwise convergent.\footnote{Kehoe \cite[Chapter 3, p.\ 22]{Kehoe:2019} stated without proof that the set of bounded-by-1 pseudometrics on any set is compact.  The proof that he provided in personal correspondence is equally applicable to bounded Euclidean pseudometrics.}
\label{lm:boundedPM}
\end{lemma}

\subparagraph{Proof (Kehoe\cite{Kehoe:2019})}
The set ${\mathcal Y}(\gamma)$ is a subset of the Cartesian product 
\[
\mbox{\tt F} = [0,\gamma]^{{\mathcal M} \times {\mathcal M}}.
\]
Because $[0,\gamma]$ is compact, it follows from Tychonoff's theorem that $\mbox{\tt F}$ is compact in the product topology, i.e., in the topology of pointwise convergence.  Because ${\mathcal Y}(\gamma)$ is closed in that topology, it must itself be compact.
\qed

\bigskip

Now we can establish a result analogous to Theorem~\ref{thm:fixed}.
\begin{theorem}
Let $n = 1,2,\ldots$ and $p \in [1,\infty)$.  Suppose that the sequence of dissimilarity functions $\{ \Delta_n \}$ is uniformly bounded and converges to the dissimilarity function $\Delta_\infty$ in the topology of pointwise convergence.
Suppose that the sequence of empirical probability measures $\{ \hat{P}_n \}$ converges weakly to the probability measure $P_\infty$.  Then 
\begin{itemize}

\item[(a)] any sequence of $D_n \in {\tt Min}(\Delta_n,\hat{P_n})$ has an accumulation point in the topology of $L^p(P_\infty)$ convergence, and 

\item[(b)] if $D_\infty$ is an accumulation point of $\{ D_n \in {\tt Min}(\Delta_n,P_n) \}$ in the topology of $L^p(P_\infty)$ convergence, then $D_\infty \in {\tt Min}(\Delta_\infty,P_\infty)$.
\end{itemize}
\label{thm:compact}
\end{theorem}

\subparagraph{Proof}
By assumption, there exists $\delta < \infty$ for which
\[
(\Delta_\infty,P_\infty),(\Delta_1,\hat{P}_1),(\Delta_2,\hat{P}_2),\ldots  \in {\mathcal X}(\delta) .
\]
By Lemma~\ref{lm:Dbounded}, each $D_n \in {\mathcal Y}(6\delta)$;
hence, by Lemma~\ref{lm:boundedPM}, each sequence of $D_n$ must contain a subsequence that converges pointwise.  

Suppose that the subsequence $\{ D_{n_k} \}$ converges pointwise to $D_\infty$.  Let 
\[
g_{n_k} \left( m_1,m_2 \right) = 
\left[ D_{n_k} \left( m_1,m_2 \right) - 
D_\infty \left( m_1,m_2 \right)
\right]^p,
\]
 which converges pointwise to zero, and note that $| g_{n_k} (m_1,m_2) | \leq (2 \cdot 6\delta)^p < \infty$ is integrable because $P_\infty$ is a finite measure.  By Lebesgue's Dominated Convergence Theorem,
\[
\lim_{k \rightarrow \infty}
\int_{{\mathcal M}} \int_{{\mathcal M}} g_{n_k} \left( m_1,m_2 \right)
P_\infty \left( dm_1 \right) P_\infty \left( dm_1 \right) = 0,
\]
establishing (a).

Claim (b) is established by arguing as in the proof of Theorem~\ref{thm:fixed}.
\qed

\bigskip

We emphasize that the importance of Theorem~\ref{thm:compact} lies in its description of the asymptotic behavior of conventional finite embedding methodology as $n$ increases.  
Even if we knew $\M$ and $(\Delta_\infty,P_\infty)$, the infinite-dimensional problem of minimizing $\sigma((\Delta_\infty,P_\infty),\cdot)$ is intractable.  Regardless, this limiting problem defines precisely what it means to extract all of the $d$-dimensional Euclidean structure latent in $\M$.  Theorem~\ref{thm:compact} establishes that any accumulation point of solutions of familiar finite-dimensional problems of minimizing $\sigma((\Delta_n,\hat{P}_n),\cdot)$ is a solution of the limiting problem, thereby providing a consistency result for standard embedding practice.

\section{Approximate Lipschitz Constraints}
\label{ApproxLipCon}

In contrast to Theorem~\ref{thm:fixed}, Theorem~\ref{thm:compact} does not establish the existence of accumulation points in the topology of uniform convergence.  This is not surprising.  Our proof of Theorem~\ref{thm:compact} relies on the Tychonoff and Dominated Convergence theorems, which do not extend from pointwise and $L^p$ convergence to $L^\infty$ convergence.  Moreover, the raw stress criterion measures $L^2$ error.  Nothing in the formulation proposed in Section~\ref{compact} leads us to expect uniform convergence.

We can, however, inquire if it is possible to obtain uniform convergence by modifying our embedding methodology.  We begin with the
Arzel\`{a}-Ascoli Theorem, which states that a sequence of continuous real-valued functions on a compact set has a subsequence that converges uniformly if and only if the sequence is uniformly bounded and uniformly equicontinuous.  If the functions have a common Lipschitz constant, then they are necessarily uniformly equicontinuous.  The challenge of the present section is to modify the problem of minimizing raw stress in such a way that resulting sequences of optimal Euclidean pseudometric functions are both uniformly bounded and uniformly Lipschitz continuous.

We begin with Lipschitz continuity.  Examining (\ref{eq:D}), it is evident that the Lipschitz continuity of a Euclidean pseudometric $D: \M \times \M \rightarrow \Re$ depends on the Lipschitz continuity of the corresponding embedding function $\mbox{\tt mds} : \M \rightarrow \Re^d$.  To encourage Lipschitz continuity of the embedding function, we impose the following {\em approximate Lipschitz constraints}:
for each $(\Delta,P)$, $D$ must satisfy
\begin{equation}
P \left( D \left( m_1,m_2 \right) =
\left\| \mbox{\tt mds}\left( m_1 \right) -
\mbox{\tt mds}\left( m_2 \right) \right\| \leq
K \Delta \left( m_1,m_2 \right) \right) = 1.
\label{eq:ALC}
\end{equation}
To explain our terminology, notice that the inequalities in (\ref{eq:ALC}) are actual Lipschitz conditions when $\Delta$ metrizes $\M$.  If $\Delta$ only approximates the metric on $\M$, then these inequalities only approximate Lipschitz conditions.

Let $\Omega_1$ denote the point-to-set map that requires $D$ to be a Euclidean pseudometric and let $\Omega_2$ denote the point-to-set map that imposes the approximate Lipschitz constraints.
Notice that each $\Omega_2(\Delta,P)$ is nonempty, if only because it necessarily includes the trivial pseudometric defined by $D(m_1,m_2)=0$.
Each constraint map is continuous, but the intersection contraint map $\Omega_K =  \Omega_1 \cap \Omega_2$ is not open.
To see this, let ${\mathcal M} = [0,1]$, let $P_n=P_\infty=\mbox{Uniform}[0,1]$, and let
$\Delta_\infty$ denote Euclidean distance.  Embed $\Delta_\infty$ in $\Re$ by $\mbox{mds}_\infty(m)=m$, with corresponding pseudometric $D_\infty=\Delta_\infty \in  \Omega_K(\Delta_\infty,P_\infty)$.  Let
\[
\Delta_n \left( m_1,m_2 \right) = \left\{ \begin{array}{ccl}
0 & \mbox{if} & \Delta_\infty \left( m_1,m_2 \right) < 1/n \\
\Delta_\infty \left( m_1,m_2 \right) & \mbox{if} & \Delta_\infty \left( m_1,m_2 \right) \geq 1/n
\end{array} \right\},
\]
and notice that $\Delta_n$ converges uniformly to $\Delta_\infty$.
If $\Omega_K$ is open, then there must be a feasible sequence $\{ D_n \in \Omega_K(\Delta_n,P_n) \}$ that converges to $D_\infty$.
But if the approximate Lipschitz constraints hold, then 
\[
\left\| \mbox{\tt mds}_n\left( m_1 \right) -
\mbox{\tt mds}_n\left( m_2 \right) \right\| \leq
K \Delta_n \left( m_1,m_2 \right) = 0
\]
for every $\| m_1-m_2 \| < 1/n$, so that $\mbox{\tt mds}_n$ must be constant on all subintervals of $[0,1]$ having length $<1/n$.  This is only possible if $\mbox{\tt mds}_n$ is constant on $[0,1]$; hence, the only feasible Euclidean pseudometric for $(\Delta_n,P_n)$ is $D_n=0$ and convergence of $\{ D_n \}$ to $D_\infty \neq 0$ is impossible.
 
It is evident from the preceding example that the approximate Lipschitz constraints are too severe if we allow $\Delta_n(m_1,m_2)=0$ when $m_1 \neq m_2$.  We cannot, however, bound $\Delta_n$ away from zero because
we must allow $\Delta_n(m_1,m_2)=0$ when $m_1=m_2$.  The difficulty is that differences are ill-suited to measuring the discrepancy between 
$\Delta_n(m_1,m_2)$ and $\Delta_\infty(m_1,m_2)$ as $\Delta_\infty(m_1,m_2) \rightarrow 0$.  Instead, we endeavor to measure discrepancy by forming ratios.

In order to control ratios of $\Delta_n$ and $\Delta_\infty$, we first modify ${\mathcal D}$, the cone of all Borel-measurable dissimilarity functions on ${\mathcal M}$.
We assume that $\Delta_\infty$ metrizes $\M$, fix $R \geq 1$, and let ${\mathcal D}_R$ consist of all continuous dissimilarity functions on ${\mathcal M}$ that satisfy
\begin{equation}
\frac{1}{R} \Delta_\infty \left( m_1,m_2 \right)  \leq
\Delta \left( m_1,m_2 \right) \leq
R \Delta_\infty \left( m_1,m_2 \right)
\label{eq:ratios}
\end{equation}
for every $m_1 \neq m_2$.
The first inequality in (\ref{eq:ratios}) ensures that $\Delta_n(m_1,m_2)$ is strictly positive when $m_1 \neq m_2$,
while the second inequality in (\ref{eq:ratios}) allows us to pass from approximate Lipschitz constraints to actual Lipschitz conditions by writing
\begin{equation}
\left\| \mbox{\tt mds}_n\left( m_1 \right) -
\mbox{\tt mds}_n\left( m_2 \right) \right\| \leq
K \Delta_n \left( m_1,m_2 \right) \leq
KR \, \Delta_\infty \left( m_1,m_2 \right).
\label{eq:KR}
\end{equation}
Furthermore, because $\M$ compact entails $\Delta_\infty \left( m_1,m_2 \right) \leq M < \infty$, we obtain 
\begin{equation}
D_n(m_1,m_2) = \left\| \mbox{\tt mds}_n\left( m_1 \right) -
\mbox{\tt mds}_n\left( m_2 \right) \right\| \leq KRM,
\label{eq:Dbounded}
\end{equation}
establishing that the $D \in \Omega(\Delta,P)$ are uniformly bounded.

In order to apply the theory of point-to-set maps, we must metrize ${\mathcal D}_R$.  In light of (\ref{eq:ratios}), a natural metric is
defined by
\[
\mu \left( \Delta_1,\Delta_2 \right) = \log \, \inf
\left\{ r : \frac{1}{r} \leq \frac{\Delta_1 \left( m_1,m_2 \right)}{\Delta_2 \left( m_1,m_2 \right)} \leq r \mbox{ for all } m_1 \neq m_2 \right\}.
\]
Equivalently, let
\[
f_i \left( m_1,m_2 \right) = \left\{ \begin{array}{ccl}
0 & \mbox{if} & m_1=m_2 \\
\log \Delta_i \left( m_1,m_2 \right) - \log \Delta_\infty \left( m_1,m_2 \right) &
\mbox{if} & m_1 \neq m_2 \end{array} \right\}.
\]
Then each $f_i$ is continuous and 
\[
\mu \left( \Delta_1,\Delta_2 \right) = \left\| f_1-f_2 \right\|_\infty,
\]
demonstrating that $({\mathcal D}_R,\mu)$ is a complete metric space.

Upon replacing ${\mathcal D}$ with ${\mathcal D}_R$, we can establish continuity of the constraint map.
\begin{lemma}
Suppose that $\Delta_\infty$ metrizes $\M$.
For any $R \geq 1$ and $p \geq 1$, the point-to-set map $\Omega_K$ is open at $(\Delta_\infty,P_\infty) \in \X = {\mathcal D}_R \times {\mathcal P}$ with respect to the topology of $L^p(P_\infty)$ convergence in $\Y$.
\label{lm:open}
\end{lemma}

\subparagraph{Proof}
Suppose that $\{ (\Delta_n,P_n) \} \subset \X$ converges to 
$(\Delta_\infty,P_\infty) \in \X$.  Choose any Euclidean pseudometric $D_\infty \in \Omega(\Y)$ and any embedding function $\mbox{\tt mds}_\infty : \M \rightarrow \Re^d$ such that 
\[
D_\infty \left( m_1,m_2 \right) = 
\left\| \mbox{\tt mds}_\infty \left( m_1 \right)-
\mbox{\tt mds}_\infty \left( m_2 \right) \right\|.
\]
Let 
\[
r_n = \inf \left\{ r : 
\frac{1}{r} \Delta_\infty \left( m_1,m_2 \right)  \leq
\Delta_n \left( m_1,m_2 \right) \leq
r \Delta_\infty \left( m_1,m_2 \right) \right\} = 
\exp \mu \left( \Delta_n,\Delta_\infty \right)
\]
and note that $r_n \rightarrow 1$ as $n \rightarrow \infty$.
Finally, set
$\mbox{\tt mds}_n (m) = \mbox{\tt mds}_\infty (m)/r_n$ and 
\[
D_n \left( m_1,m_2 \right) =
\left\| \mbox{\tt mds}_n \left( m_1 \right)-
\mbox{\tt mds}_n \left( m_2 \right) \right\| \\
 =  \left\| \mbox{\tt mds}_\infty \left( m_1 \right)-
\mbox{\tt mds}_\infty \left( m_2 \right) \right\| / r_n.
\]
As $n \rightarrow \infty$, $D_n \rightarrow D_\infty$ pointwise; hence, by Lebesgue's Dominated Convergence Theorem, in $L^p(P_\infty)$.
Finally, 
\begin{eqnarray*}
D_n \left( m_1,m_2 \right) & = &
\left\| \mbox{\tt mds}_\infty \left( m_1 \right)-
\mbox{\tt mds}_\infty \left( m_2 \right) \right\| / r_n \\
 & \leq & K \Delta_\infty \left( m_1,m_2 \right) / r_n \\
  & \leq & K r_n \, \Delta_n \left( m_1,m_2 \right) / r_n \\
 & = & K \Delta_n \left( m_1,m_2 \right),
\end{eqnarray*}
i.e., $D_n \in \Omega(\Delta_n,P_n)$.
\qed

\bigskip

Now we can study embedding problems of the form
\begin{equation}
\begin{array}{ll}
\mbox{minimize} & \sigma \left( (\Delta, P), D \right) =
\left\| D-\Delta \right\|_P^2 \\ 
\mbox{subject to} & D \in \Omega_K(\Delta,P).
\end{array}
\label{pr:ALE}
\end{equation} 
Solving (\ref{pr:ALE}) when $P=\hat{P}_n$, the empirical distribution of $m_1,\ldots,m_n \in \M$, results in a new technique for traditional MDS, {\em Approximate Lipschitz Embedding}\/ (ALE).  We describe a computational algorithm for ALE in the Appendix.

Just as Theorem~\ref{thm:compact} establishes asymptotic consistency  for unconstrained minimization of raw stress, the following result establishes asymptotic consistency for ALE.
\begin{theorem}
Let $n = 1,2,\ldots$, $p \in [1,\infty)$, and $R \geq 1$.  Suppose that  $\Delta_\infty$ metrizes $\M$ and that the sequence of dissimilarity functions $\{ \Delta_n \in {\mathcal D}_R \}$ converges to $\Delta_\infty$ in the topology of the metric $\mu$.
Suppose that the sequence of empirical probability measures $\{ \hat{P}_n \}$ converges weakly to the probability measure $P_\infty$.  Then 
\begin{itemize}

\item[(a)] any sequence of $D_n \in {\tt Min}(\Delta_n,\hat{P_n})$ has an accumulation point in the topology of $L^p(P_\infty)$ convergence, and 

\item[(b)] if $D_\infty$ is an accumulation point of $\{ D_n \in {\tt Min}(\Delta_n,\hat{P}_n) \}$ in the topology of $L^p(P_\infty)$ convergence, then $D_\infty \in {\tt Min}(\Delta_\infty,P_\infty)$.
\end{itemize}
\label{thm:ALE}
\end{theorem}

\subparagraph{Proof}
The proof is identical to the proof of Theorem~\ref{thm:compact}, except that we rely on (\ref{eq:Dbounded}) instead of Lemma~\ref{lm:Dbounded} to establish the uniform boundedness of the sequence of pseudometrics $\{ D_n \}$.
\qed

\bigskip

Although the purpose of imposing approximate Lipschitz constraints was to encourage Lipschitz continuity and thus obtain uniform convergence, Theorem~\ref{thm:ALE} only establishes asymptotic consistency of the optimal pseudometrics in the sense of $L^p$ convergence.  It remains to determine whether or not a sequence of optimal pseudometrics converges uniformly.  

Notice that embedding $(\Delta_n,\hat{P}_n)$ only specifies a solution on $\{ m_1,\ldots,m_n \}$.  Let $z_1,\ldots,z_n \in \Re^d$ be any optimal configuration of points constructed by ALE.  Any embedding function $\mbox{\tt mds}_n : {\mathcal M} \rightarrow \Re^d$ that interpolates the $z_i$ defines an optimal pseudometric $D_n$.  If we construct $\{ \mbox{\tt mds}_n \}$ with interpolating functions that oscillate more and more wildly as $n$ increases, then it is clear that we can preclude uniform Lipschitz continuity of $\{ \mbox{\tt mds}_n \}$ and uniform convergence of the corresponding $\{ D_n \}$.  We now proceed to demonstrate that it is also possible to interpolate $z_1,\ldots,z_n$ in a way that guarantees uniform Lipschitz continuity of $\{ \mbox{\tt mds}_n \}$ and (therefore) uniform convergence of the corresponding $\{ D_n \}$.

We require several technical lemmas about Lipschitz continuity.  The first extends a result from approximation theory, from $f : \Re^m \rightarrow \Re$ to $f : \M \rightarrow \Re$.  
\begin{lemma}[Beliakov{\cite[Theorem~4]{Beliakov:2006}}]
Suppose that $\Delta_\infty$ metrizes $\M$, 
and that \\ $m_1,\ldots,m_n \in \M$ and $z_1,\ldots,z_n \in \Re$ satisfy
\[
\left| z_i-z_j \right| \leq c \, \Delta_\infty \left( m_1,m_2 \right)
\]
for every $1 \leq i,j \leq n$.
Then there exists a function $f : \M \rightarrow \Re$ such that $f(m_i) = z_i$, $i=1,\ldots,n$, and
\[
\left| f(m)-f(m^\prime) \right| \leq c \, \Delta_\infty \left( m,m^\prime \right)
\]
for every $m,m^\prime \in \M$.
\label{lm:Beliakov}
\end{lemma}

\subparagraph{Proof (Beliakov\cite{Beliakov:2006})}
For $k=1,\ldots,n$, define $h_k : \M \rightarrow \Re$ by
\[
h_k(m) = z_k - c \, \Delta_\infty \left( m,m_k \right).
\]
Notice that $h_k(m_k)=z_k$, and that
\[
\left| h_k (m) - h_k \left( m^\prime \right) \right| =
\left| -c \, \Delta_\infty \left( m,m_k \right) +
c \, \Delta_\infty \left( m^\prime,m_k \right) \right|
\leq
c \, \Delta_\infty \left( m,m^\prime \right)
\]
by the reverse triangle inequality.  Now define $H : \M \rightarrow \Re$ to be the pointwise maximum of the $h_k$, i.e.,
$H(m) = \max_{k=1,\ldots,n} h_k(m)$.  Then
\[
H \left( m_k \right) = h_k \left( m_k \right) = z_k,
\]
and
\begin{eqnarray*}
\left| H(m) - H \left( m^\prime \right) \right| & = &
\left| \max_k h_k(m) - \max_k h_k \left( m^\prime \right) \right| \\
& \leq &
\max_k \left| h_k(m) - h_k \left( m^\prime \right) \right| \leq
c \, \Delta_\infty \left( m,m^\prime \right).
\end{eqnarray*}
\qed

\bigskip

Next we note that Lipschitz continuity of a vector-valued function is implied by Lipschitz continuity of its scalar-valued component functions.
\begin{lemma}
Suppose that $\Delta_\infty$ metrizes $\M$, and that $f_1,\ldots,f_d : \M \rightarrow \Re$ each satisfies
\[
\left| f_i \left( m_1 \right) - f_i \left( m_2 \right) \right| \leq
c \, \Delta_\infty \left( m_1,m_2 \right)
\]
for every $m_1,m_2 \in \M$.  Then $F : \M \rightarrow \Re^d$ defined by $F(m) = [ \begin{array}{c|c|c} f_1(m) & \cdots & f_d(m) \end{array} ]^t$ satisfies
\[
\left\| F \left( m_1 \right) - F \left( m_2 \right) \right\| \leq
c\sqrt{d} \, \Delta_\infty \left( m_1,m_2 \right)
\]
for every $m_1,m_2 \in \M$.
\label{lm:vectorLipCon}
\end{lemma}

\subparagraph{Proof}
\begin{eqnarray*}
\left\| F \left( m_1 \right) - F \left( m_2 \right) \right\|^2 & = & 
\sum_{i=1}^d \left| f_i \left( m_1 \right) - f_i \left( m_2 \right) \right|^2 \\ & \leq &
\sum_{i=1}^d c^2 \Delta_\infty^2 \left( m_1,m_2 \right) =
dc^2 \Delta_\infty^2 \left( m_1,m_2 \right)
\end{eqnarray*}
\qed

\bigskip

Finally, if $\Delta_\infty$ metrizes $\M$,
then the squared product distance between $m=(m_1,m_2) \in \M \times \M$ and $m^\prime=(m^\prime_1,m^\prime_2) \in \M \times \M$ is
\[
\Delta_\infty^2 \left( m_1,m^\prime_1 \right) + \Delta_\infty^2 \left( m_2,m^\prime_2 \right).
\]
The following lemma establishes that,
if the embedding function $\mbox{\tt mds} : \M \rightarrow \Re^d$ is Lipschitz continuous with respect to $\Delta_\infty$,
then the Euclidean pseudometric $D : \M \times \M \rightarrow \Re$ is Lipschitz continuous with respect to the corresponding product metric on $\M \times \M$.
\begin{lemma}
Suppose that $\Delta_\infty$ metrizes $\M$.  If
\[
D \left( m_1,m_2 \right) =
\left\| \mbox{\tt mds} \left( m_1 \right) - 
\mbox{\tt mds} \left( m_2 \right) \right\| \leq
c \, \Delta_\infty \left( m_1,m_2 \right)
\]
for every $m_1,m_2 \in \M$, then
\[
\left| D \left( m_1,m_2 \right) - D \left( m^\prime_1,m^\prime_2 \right) \right| \leq
c\sqrt{3} \, \left[ \Delta_\infty^2 \left( m_1,m^\prime_1 \right) + \Delta_\infty^2 \left( m_2,m^\prime_2 \right) \right]^{1/2}
\]
for every $m_1,m_2,m^\prime_1,m^\prime_2 \in \M$.
\label{lm:DLipCon}
\end{lemma}

\subparagraph{Proof}
For any $a,b \geq 0$,
\[
ab \leq \max (a^2,b^2) \leq a^2+b^2,
\]
hence
\[
(a+b)^2 = a^2+b^2 + 2ab \leq 3(a^2+b^2).
\]
It follows from the triangle inequality that
\[
D \left( m_1,m_2 \right) \leq 
D \left( m_1,m^\prime_1 \right) +
D \left( m^\prime_1,m^\prime_2 \right) +
D \left( m^\prime_2,m_2 \right)
\]
and
\[
D \left( m^\prime_1,m^\prime_2 \right) \leq 
D \left( m^\prime_1,m_1 \right) +
D \left( m_1,m_2 \right) +
D \left( m_2,m^\prime_2 \right),
\]
from which we obtain
\begin{eqnarray*}
\left| D \left( m_1,m_2 \right) - D \left( m^\prime_1,m^\prime_2 \right) \right|
 & \leq & D \left( m_1,m^\prime_1 \right) + D \left( m_2,m^\prime_2 \right) \\ 
 & \leq & c \, \Delta_\infty \left( m_1,m^\prime_1 \right) + 
 c \, \Delta_\infty \left( m_2,m^\prime_2 \right) \\ & \leq &
c \sqrt{3} \left[ \Delta_\infty^2 \left( m_1,m^\prime_1 \right) + 
\Delta_\infty^2 \left( m_2,m^\prime_2 \right) \right]^{1/2}.
\end{eqnarray*}
\qed

\bigskip

We can now strengthen the mode of convergence in Theorem~\ref{thm:ALE}.
\begin{theorem}
Let $n = 1,2,\ldots$, $p \in [1,\infty)$, and $R \geq 1$.  Suppose that  $\Delta_\infty$ metrizes $\M$ and that the sequence of dissimilarity functions $\{ \Delta_n \in {\mathcal D}_R \}$ converges to $\Delta_\infty$ in the topology of the metric $\mu$.
Suppose that the sequence of empirical probability measures $\{ \hat{P}_n \}$ converges weakly to the probability measure $P_\infty$.  
Then, for any sequence of $D_n \in {\tt Min}(\Delta_n,\hat{P_n})$,
there exists a corresponding sequence $\{ \bar{D}_n \}$ with the following properties:
\begin{itemize}

\item[(a)]  If $m_1,m_2 \in \M$ lie in the support of $\hat{P}_n$, then $\bar{D}_n(m_1,m_2) = D_n(m_1,m_2)$.

\item[(b)] Each $\bar{D}_n \in {\tt Min}(\Delta_n,\hat{P_n})$.

\item[(c)] The sequence $\{ \bar{D}_n \}$ has an accumulation point in the topology of uniform convergence, and 

\item[(d)] if $\bar{D}_\infty$ is an accumulation point of $\{ \bar{D}_n \}$ in the topology of uniform convergence, then $\bar{D}_\infty \in {\tt Min}(\Delta_\infty,P_\infty)$.
\end{itemize}
\label{thm:uniformALE}
\end{theorem}

\subparagraph{Proof}
From (\ref{eq:KR}), we have
\[
D_n \left( m_1,m_2 \right) =
\left\| \mbox{\tt mds}_n \left( m_1 \right) -
\mbox{\tt mds}_n \left( m_2 \right) \right\| \leq
KR \, \Delta_\infty \left( m_1,m_2 \right).
\]
For each component function of $\mbox{\tt mds}_n : \M \rightarrow \Re^d$, Lemma~\ref{lm:Beliakov} guarantees the existence of $f_i : \M \rightarrow \Re$ that satisfies 
\begin{center}
$f_i(m_k)=\mbox{\tt mds}_n(m)$ for each $m$ in the support of $\hat{P}_n$,
\end{center}
and, upon setting $c=KR$,
\[
\left| f_i \left( m_1 \right) -
f_i \left( m_2 \right) \right| \leq
KR \, \Delta_\infty \left( m_1,m_2 \right).
\]
Let $F_n : \M \rightarrow \Re^d$ denote the embedding function whose component functions are $f_1,\ldots,f_d$.  We then obtain, from Lemma~\ref{lm:vectorLipCon} with $c=KR$,
\[
\left\| F_n \left( m_1 \right) -
F_n \left( m_2 \right) \right\| \leq
KR\sqrt{d} \, \Delta_\infty \left( m_1,m_2 \right).
\]
Finally, define $\bar{D}_n : \M \times \M \rightarrow \Re$ by
\[
\bar{D}_n \left( m_1,m_2 \right) =
\left\| F_n \left( m_1 \right) -
F_n \left( m_2 \right) \right\|
\]
and apply Lemma~\ref{lm:DLipCon} with $c=KR\sqrt{d}$ to obtain
\begin{equation}
\left| \bar{D}_n \left( m_1,m_2 \right) - \bar{D}_n \left( m^\prime_1,m^\prime_2 \right) \right| \leq
KR\sqrt{3d} \, \left[ \Delta_\infty^2 \left( m_1,m^\prime_1 \right) + \Delta_\infty^2 \left( m_2,m^\prime_2 \right) \right]^{1/2}.
\label{eq:barD}
\end{equation}

Property (a) is evident from the construction of the $f_i$, and property (b) follows immediately because $\sigma((\Delta_n,\hat{P}_n),D))$ only considers pairs of points that lie in the support of $\hat{P}_n$.
The sequence $\{ \bar{D}_n \}$ is uniformly bounded in light of (\ref{eq:Dbounded}), and
(\ref{eq:barD}) demonstrates that it is uniformly Lipschitz continuous.
Property (c) then follows from the Arzel\`{a}-Ascoli Theorem.
Because uniform convergence entails $L^p(P_\infty)$ convergence,
property (d) then follows from part (b) of Theorem~\ref{thm:ALE}.
\qed

\bigskip

We remark that Theorem~\ref{thm:uniformALE} is not constructive in any practical sense, as Lemma~\ref{lm:Beliakov} uses $\Delta_\infty$ to construct the required interpolating functions.  In practice, $\Delta_\infty$ is unknown.  We defer to future investigation the question of whether practical interpolation methods, e.g., interpolation by radial basis functions, can be constructed so that they possess the desired Lipschitz continuity properties.

\section{Discussion}
\label{disc}

Kruskal's \cite{kruskal:1964a} raw stress criterion for MDS is actually not a single objective function, but a family of formally distinct objective functions that vary with the number of objects to be embedded.  We have formulated a mathematical framework for {\em continuous MDS}\/ in which all problems that minimize a raw stress criterion can be realized as special cases.  The continuous embedding problems are mathematically intractable, but easily interpreted as limiting cases of the traditional embedding problems.  Within this framework, we have derived consistency results for the asymptotic behavior of solutions to both unconstrained and constrained embedding problems that minimize raw stress.

Although we believe that our formulation of continuous MDS in Section~\ref{compact} is of general interest, the work described herein was motivated by our study of random dot product graphs (RDPGs).  See \cite{mwt:rdpg1} and the references therein.
Briefly, consider an RDPG with compact Riemannian support manifold $\M$.  Draw $n$ latent positions on $\M$, then generate an adjacency matrix.  To draw inferences about an unknown $\M$, we have proposed the following methodology\cite{mwt:rdpg1}:
\begin{enumerate}

\item  Estimate latent positions $V \subset \Re^k$ by adjacency spectral embedding.

\item  Learn $\M$ by Isomap:
\begin{enumerate}

\item  Approximate Riemannian distance on $\M$ by shortest path distance on a suitable graph ${\mathcal G}=(V,E)$ that localizes the structure of $\M$.

\item  Approximate the shortest path distances by Euclidean distances in $\Re^d$.

\end{enumerate}

\item Construct a decision rule in $\Re^d$.

\end{enumerate}
This methodology exemplifies {\em manifold learning for subsequent inference}.  If $\M$ is known, then {\em restricted inference}\/ derives a decision rule that exploits $\M$.  If $\M$ is unknown, then we hope to learn enough about $\M$ to realize some benefit of restricted inference.  In particular, does the performance of rules based on the learnt manifold converge to the performance of a rule based on $\M$ as $n \rightarrow \infty$?

To answer such questions, one must understand how the approximating Euclidean distances in 2(b) behave as $n \rightarrow \infty$.  The analysis in \cite[Section~5.5]{mwt:rdpg1} assumes that $\M$ is $1$-dimensional and that out-of-sample embedding is used for $n$ sufficiently large.  Theorem~\ref{thm:compact} improves on that analysis, allowing $d>1$ and dispensing with out-of-sample embedding.

Of course, stronger modes of convergence for the embedding procedures that construct data representations in $\Re^d$ will allow us to establish stronger results for the asymptotic behavior of decision rules based on those representations.
This is the reason that we sought to strengthen the mode of convergence from $L^p$ to $L^\infty$ in Section~\ref{ApproxLipCon}.  There we demonstrated that uniform convergence can be obtained by minimizing the raw stress criterion subject to {\em approximate Lipschitz constraints}, a new method of multidimensional scaling that we have named {\em Approximate Lipschitz Embedding}\/ (ALE).

The methods and results in Section~\ref{ApproxLipCon} invite (at least) two future lines of research.  First, in order to realize the asymptotic behavior described by our convergence results, we need to develop efficient algorithms for performing ALE when $n$ is large.  The algorithm described in the Appendix can surely be improved.  Second,
in order to obtain performance guarantees, e.g., concentration inequalities and rates of convergence, for subsequent inference, we will likely need to derive Lipschitz constants for specific ways of interpolating configurations constructed by ALE.  Interpolation techniques that are based on distances in the domain of the function to be interpolated, e.g., radial basis functions and inverse distance weighting, appear to be especially well-suited to our needs.

\section*{Appendix: Computation}
\label{compute}

\subsection*{Unconstrained Minimization of Raw Stress}

Let $\Delta$ denote a fixed $n \times n$ dissimilarity matrix and consider the unconstrained optimization problem of finding an $n \times d$ configuration matrix $Z$ that minimizes (\ref{eq:rawstress}).  A standard approach to unconstrained optimization begins with identifying solutions of the stationary equation, i.e., the equation that requires all partial derivatives of the objective function to vanish.  A popular approach to minimizing (\ref{eq:rawstress}) exploits the special structure of its stationary equation.

Let $\G$ denote an undirected graph with vertices $1,\ldots,n$.
If $w_{ij}=w_{ji}>0$, then vertices $i$ and $j$ are connected with
edge weight $w_{ij}$.  Assume that $\G$ is connected, in which case $L$, the combinatorial Laplacian matrix of $\G$, has exactly one zero eigenvalue with eigenvector $e=(1,\ldots,1) \in \Re^n$.
Next, modify $\G$ by removing any edges that connect identical vertices, i.e., vertices $i$ and $j$ for which $d_{ij} = \| z_i-z_j \| =0$.
Assign edge weights of $w_{ij} \delta_{ij}/d_{ij}$ to the remaining edges and let $M(Z)$ denote the combinatorial Laplacian matrix of the modified graph.  Then it turns out that the stationary equation for minimizing (\ref{eq:rawstress}) can be written as $LZ = M(Z)Z$.

The equation $LZ = M(Z)Z$ suggests an iterative method for finding stationary configurations: given a configuration $Z_k$, solve the linear system $LZ = M(Z_k)Z_k$ to obtain a new configuration $Z_{k+1}$.  Typically, one computes the unique solution that satisfies $e^tZ = \vec{0} \in \Re^d$, i.e., one requires the configuration $Z_{k+1}$ to be centered at the origin of $\Re^d$.  This solution can be written as
\begin{equation}
Z_{k+1} = \Gamma \left( Z_k \right) =
L^\dagger M \left( Z_k \right) Z_k =
\left( L + ee^t \right)^{-1} M \left( Z_k \right) Z_k,
\label{eq:guttman}
\end{equation}
and is commonly known as the {\em Guttman transform}\/ of $Z_k$.
It is well-known that, if $Z_k$ is not a stationary configuration, then
\[
\sigma_n \left( \Delta,Z_{k+1} \right) < 
\sigma_n \left( \Delta,Z_k \right),
\]
and furthermore that any sequence $\{ \sigma_n \left( \Delta,Z_k \right) \}$ converges to the raw stress value of a stationary configuration.  See, for example, the convergence analysis in \cite{deleeuw:1988}.  Because the sequence of raw stress values is nonincreasing, it is highly improbable that this stationary configuration will be anything other than a local minimizer of (\ref{eq:rawstress}).

Notice that computing (\ref{eq:guttman}) requires an initial matrix factorization, which may be expensive if $n$ is large.  Fortunately, there are special cases---including the case of equal $w_{ij}$---
in which (\ref{eq:guttman}) simplifies, obviating matrix factorization.

The limitations of the above method are also well-known.
Typically, (\ref{eq:rawstress}) will have nonglobal minimizers.
Because the method of Guttman transforms cannot distinguish global minimizers from nonglobal minimizers, it is strongly recommended that one start from a good initial configuration.  At least when the $w_{ij}$ are equal, setting $Z_1$ equal to the configuration constructed by classical MDS typically produces good results.  Because classical MDS also requires a matrix factorization, one may prefer less expensive ways of constructing initial configurations, e.g., the landmark MDS technique described in \cite{desilva&tenenbaum:2004}.

It is also known that (\ref{eq:rawstress}) does not have much curvature near solutions.  Because the Guttman method is based entirely on the first-order stationary equation, it does not incorporate second-order information about (\ref{eq:rawstress}) and therefore converges quite slowly.  Indeed, one can think of the Guttman method as a weighted gradient algorithm in which the Guttman transform identifies both a descent direction and the length of the step to take in that direction.  If precise solutions are desired, then one may prefer a hybrid method: start with the Guttman method, taking advantage of its excellent global convergence properties, then switch to the globalized Newton's method described in \cite{kearsley&etal:newton}, taking advantage of its fast local convergence properties.

\subsection*{Constrained Minimization of Raw Stress}

Enforcing the approximate Lipschitz constraints in the case of $(\Delta,\hat{P}_n)$, where $\hat{P}_n$ is the empirical distribution of $m_1,\ldots,m_n \in \M$, results in a new technique for traditional MDS.  Let $\delta_{ij} = \Delta(m_i,m_j)$ and $\Delta_n = [ \delta_{ij} ]$.
Stated in the conventional MDS formulation in which the decision variables are $n \times p$ configuration matrices,
{\em approximate Lipschitz embedding}\/ (ALE)
is defined by the constrained optimization problem
\[
\begin{array}{ll}
\mbox{minimize} & \sigma_n(\Delta_n,Z) =  \sum_{i,j=1}^n w_{ij} \left[ 
\left\| z_i-z_j \right\| - \delta_{ij} \right]^2 \\*[10pt]
\mbox{subject to} & \left\| z_i-z_j \right\| \leq K \delta_{ij}
\mbox{ for } 1 \leq i < j \leq n.
\end{array}
\label{pr:ALE}
\]
This is an example of MDS with restrictions on the configuration.

The feasible set of approximate Lipschitz constraint $ij$ in (\ref{pr:ALE}) is obviously closed.  Furthermore,
suppose that $X$ and $Y$ are two $n \times d$ configuration matrices that satisfy constraint $ij$.  Then, for $t \in [0,1]$,
\begin{eqnarray*}
\left\| \left[ tx_i+(1-t)y_i \right] - \left[ tx_j+(1-t)y_j \right] \right\|^2 & = &
\left\| t \left[ x_i - x_j \right] + (1-t) \left[ y_i - y_j \right] \right\|^2 \\ & \leq &
t \left\| x_i - x_j \right\|^2 + (1-t) \left\| y_i - y_j \right\|^2 \\ & \leq &
K^2 \delta_{ij}^2,
\end{eqnarray*}
and it follows that the feasible set of constraint $ij$ is also convex.
The feasible set for (\ref{pr:ALE}) is thus the intersection of closed and convex sets, hence closed and convex itself.

A projected Guttman method for minimizing the raw stress criterion subject to closed and convex restrictions on the configuration was analyzed in \cite{deleeuw&heiser:1980}.
Let $\Omega$ denote the closed and convex feasible set of configurations and let $P_\Omega$ denote projection into $\Omega$.  Then
\[
Z_{k+1} = P_\Omega \left( \Gamma \left( Z_k \right) \right),
\]
which simplifies to the usual Guttman method in the absence of restrictions.  Their properties are analogous, except that convergence statements involve constrained stationary configurations rather than stationary configurations.  In particular, the following results were established:
\begin{theorem}[Theorems 7 \& 8 in \cite{deleeuw&heiser:1980}]
Let $\{ Z_k \}$ be a sequence of configurations obtained by iterations of the projected Guttman method.  Then
\begin{itemize}

\item[7.]  The sequence $\{ Z_k \}$ has convergent subsequences.  Each subsequential limit is a constrained stationary point.  All subsequential limits have the same value of $\sigma_n$.

\item[8.]  The quantity
\[
\mbox{\tt trace} \left( Z_{k+1}-Z_k \right)^t L \left( Z_{k+1}-Z_k \right)^t
\] 
converges to zero.

\end{itemize}
\end{theorem}

For ALE, $\Omega$ denotes the approximate Lipschitz constraints.  To compute an approximate Lipschitz embedding by the projected Guttman method, one must be able to compute projections into $\Omega$.
Notice that projection into constraint $ij$ is trivial: if $Z$ is infeasible, then move $z_i$ and $z_j$ closer together along the line segment that connects them until constraint $ij$ is satisfied.
In consequence, projections into $\Omega$ can be computed by Dykstra's cyclic projection algorithm \cite{GaffkeMathar:1989}.

At least in theory, then, one can implement ALE as readily as one can minimize the unconstrained raw stress criterion.  What is presently unclear is the computational expense of managing ${n \choose 2}$ approximate Lipschitz constraints.  The cost of projecting into a single constraint is negligible, but the number of Lipschitz constraints is $O(n^2)$.  Presumably, one can develop more efficient algorithms that settle for approximate projections when $k$ is small and improve the accuracy of projection as $k$ increases.  The question of how many objects can be managed by an efficient implementation of ALE is a topic for future research.

\section*{Acknowledgments}
This work was partially supported by the Naval Engineering Education Consortium (NEEC), Office of Naval Research (ONR) Award Number N00174-19-1-0011; ONR Award Number N00024-22-D-6404 (via Johns Hopkins University APL); and ONR Award Number N00014-24-1-2278 (Science of Autonomy).
The first author benefitted enormously from discussions with G\"{o}k\c{c}en B\"{u}y\"{u}kba\c{s}.
Eric Kehoe generously provided the proof of Lemma~\ref{lm:boundedPM}.

\bibliography{$HOME/lib/tex/stat,$HOME/lib/tex/mds,$HOME/lib/tex/math,$HOME/lib/tex/mwt,$HOME/lib/tex/cep,$HOME/lib/tex/bio,$HOME/lib/tex/proximity,$HOME/lib/tex/na1,$HOME/lib/tex/na2}

\begin{thebibliography}{18}
\providecommand{\natexlab}[1]{#1}
\providecommand{\url}[1]{\texttt{#1}}
\expandafter\ifx\csname urlstyle\endcsname\relax
  \providecommand{\doi}[1]{doi: #1}\else
  \providecommand{\doi}{doi: \begingroup \urlstyle{rm}\Url}\fi

\bibitem[Anisiu(1981)]{Anisiu:1981}
M.-C. Anisiu.
\newblock Point-to-set mappings. {C}ontinuity.
\newblock Preprint~3, Faculty of Mathematics, Babes-Bolyai University, 1981.
\newblock Research Seminar on Fixed Point Theory.

\bibitem[Beliakov(2006)]{Beliakov:2006}
G.~Beliakov.
\newblock Interpolation of {L}ipschitz functions.
\newblock \emph{Journal of Computational \& Applied Mathematics}, 196:\penalty0
  20--44, 2006.

\bibitem[Bernstein et~al.(2000)Bernstein, de~Silva, Langford, and
  Tenenbaum]{Bernstein&etal:2000}
M.~Bernstein, V.~de~Silva, J.~C. Langford, and J.~B. Tenenbaum.
\newblock Graph approximations to geodesics on embedded manifolds. \\
\newblock \verb+https://web.mit.edu/cocosci/isomap/BdSLT.pdf+, December 20,
  2000.

\bibitem[de~Leeuw(1988)]{deleeuw:1988}
J.~de~Leeuw.
\newblock Convergence of the majorization method for multidimensional scaling.
\newblock \emph{Journal of Classification}, 5:\penalty0 163--180, 1988.

\bibitem[de~Leeuw and Heiser(1980)]{deleeuw&heiser:1980}
J.~de~Leeuw and W.~Heiser.
\newblock Multidimensional scaling with restrictions on the configuration.
\newblock In P.~R. Krishnaiah, editor, \emph{Multivariate Analysis}, volume~5,
  pages 501--522. North-Holland Publishing Company, Amsterdam, 1980.

\bibitem[de~Silva and Tenenbaum(2004)]{desilva&tenenbaum:2004}
V.~de~Silva and J.~B. Tenenbaum.
\newblock Sparse multidimensional scaling using landmark points.
\newblock Available at \\
  {\verb+http://mypage.iu.edu/~mtrosset/Courses/675/LMDS2004.pdf+}, June 2004.

\bibitem[Gaffke and Mathar(1989)]{GaffkeMathar:1989}
N.~Gaffke and R.~Mathar.
\newblock A cyclic projection algorithm via duality.
\newblock \emph{Metrika}, 36:\penalty0 29--54, 1989.

\bibitem[Hogan(1973)]{Hogan:1973}
W.~W. Hogan.
\newblock Point-to-set maps in mathematical programming.
\newblock \emph{{SIAM} Review}, 15\penalty0 (3):\penalty0 591--603, 1973.

\bibitem[Kearsley et~al.(1998)Kearsley, Tapia, and
  Trosset]{kearsley&etal:newton}
A.~J. Kearsley, R.~A. Tapia, and M.~W. Trosset.
\newblock The solution of the metric {STRESS} and {SSTRESS} problems in
  multidimensional scaling using {N}ewton's method.
\newblock \emph{Computational Sta\-tis\-tics}, 13\penalty0 (3):\penalty0
  369--396, 1998.

\bibitem[Kehoe(2019)]{Kehoe:2019}
E.~R. Kehoe.
\newblock \emph{Pseudometrics, the Complex of Ultrametrics, and Iterated Cycle
  Structures}.
\newblock PhD thesis, University of New Hampshire, Durham, 2019.

\bibitem[Kruskal(1964)]{kruskal:1964a}
J.~B. Kruskal.
\newblock Multidimensional scaling by optimizing goodness of fit to a nonmetric
  hypothesis.
\newblock \emph{Psychometrika}, 29:\penalty0 1--27, 1964.

\bibitem[Malone et~al.(2002)Malone, Tarazaga, and Trosset]{malone&etal:start}
S.~W. Malone, P.~Tarazaga, and M.~W. Trosset.
\newblock Better initial configurations for metric multidimensional scaling.
\newblock \emph{Computational Statistics and Data Analysis}, 41\penalty0
  (1):\penalty0 143--156, 2002.

\bibitem[Tenenbaum et~al.(2000)Tenenbaum, de~Silva, and Langford]{isomap:2000}
J.~B. Tenenbaum, V.~de~Silva, and J.~C. Langford.
\newblock A global geometric framework for nonlinear dimensionality reduction.
\newblock \emph{Science}, 290:\penalty0 2319--2323, 2000.

\bibitem[Torgerson(1952)]{torgerson:1952}
W.~S. Torgerson.
\newblock Multidimensional scaling: {I}. {T}heory and method.
\newblock \emph{Psychometrika}, 17:\penalty0 401--419, 1952.

\bibitem[Trosset and B\"{u}y\"{u}kba{\c{s}}(2020)]{mwt:isomap}
M.~W. Trosset and G.~B\"{u}y\"{u}kba{\c{s}}.
\newblock Rehabilitating {I}somap: {E}uclidean representation of geodesic
  structure.
\newblock arXiv:2006.10858, 2020.

\bibitem[Trosset et~al.(2020)Trosset, Gao, Tang, and Priebe]{mwt:rdpg1}
M.~W. Trosset, M.~Gao, M.~Tang, and C.~E. Priebe.
\newblock Learning $1$-dimensional submanifolds for subsequent inference on
  random dot product graphs.
\newblock arXiv:2004.07348, 2020.

\bibitem[van Gaans(Winter 2002--2003)]{GaansProbabilityMO}
O.~van Gaans.
\newblock Probability measures on metric spaces. \\
\newblock \verb+https://www.math.leidenuniv.nl/~vangaans/jancol1.pdf+, Winter
  2002--2003.

\bibitem[Zangwill(1969)]{Zangwill:1969}
W.~I. Zangwill.
\newblock \emph{Nonlinear Programming: A Unified Approach}.
\newblock Prentice-Hall, Englewood Cliffs, NJ, 1969.

\end{thebibliography}

\end{document}